\documentclass{opendatalab}

\usepackage[utf8]{inputenc}
\usepackage[most]{tcolorbox}
\usepackage{longtable}
\usepackage{listings}
\usepackage[numbers]{natbib}
\usepackage{enumitem}
\usepackage{graphicx}
\usepackage{xspace} 
\usepackage{amsmath}
\usepackage{algorithm}
\usepackage{algpseudocode} 
\usepackage{titletoc}
\usepackage{subcaption}
\usepackage[table]{xcolor}
\definecolor{lightblue}{rgb}{0.93,0.95,1.0}
\usepackage{tabularx}
\usepackage{wrapfig}
\usepackage{multirow}
\usepackage[T1]{fontenc}
\usepackage{hyperref} 
\usepackage{url}
\usepackage{booktabs} 
\usepackage{amsfonts} 
\usepackage{nicefrac} 
\usepackage{microtype}

\definecolor{codegreen}{rgb}{0,0.6,0}
\definecolor{codegray}{rgb}{0.5,0.5,0.5}
\definecolor{codepurple}{rgb}{0.58,0,0.82}
\definecolor{backcolour}{rgb}{0.95,0.95,0.92}
\definecolor{promptcolor}{HTML}{D1D0F2}
\definecolor{promptcolorheader}{HTML}{bdbcec}

\newcommand{\web}{\raisebox{-1.5pt}{\includegraphics[height=1.05em]{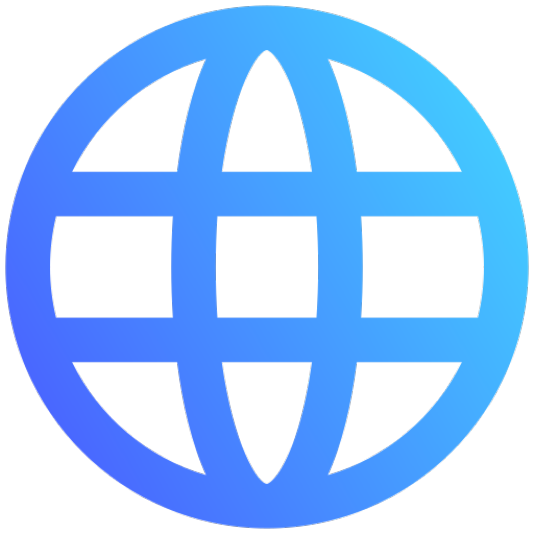}}\xspace}
\newcommand{\huggingface}{\raisebox{-1.5pt}{\includegraphics[height=1.05em]{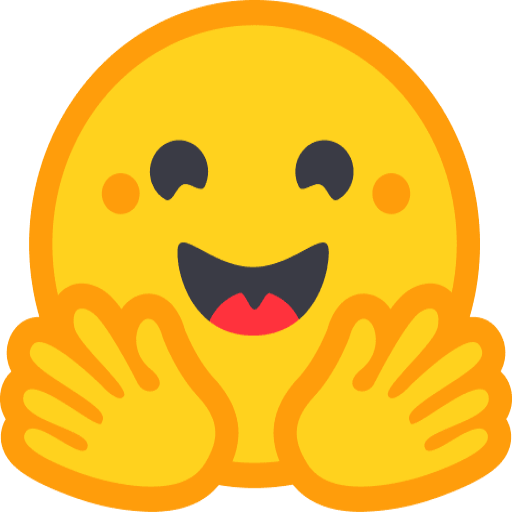}}\xspace}

\definecolor{promptcolor}{HTML}{E3F0FA}
\definecolor{promptcolorheader}{HTML}{B5D6ED}
\definecolor{prompttitletext}{HTML}{1B3A5C}

\newtcolorbox{promptbox}[1][]{
  enhanced, breakable,
  top=0.3em,bottom=0.3em,left=0.5em,right=0.5em,
  toptitle=0.3em,bottomtitle=0.2em,boxsep=0pt,
  colframe=promptcolorheader, colback=promptcolor!50, boxrule=0.5pt,
  width=\columnwidth, 
  coltitle=prompttitletext,
  title={\footnotesize #1} 
}
\lstdefinestyle{promptstyle}{
    backgroundcolor=\color{backcolour},   
    commentstyle=\color{codegreen},
    keywordstyle=\color{magenta},
    numberstyle=\tiny\color{codegray},
    stringstyle=\color{codepurple},
    basicstyle=\ttfamily\footnotesize,
    breakatwhitespace=false,         
    breaklines=true,                 
    captionpos=b,                    
    keepspaces=true,                 
    numbers=left,                    
    numbersep=5pt,                  
    showspaces=false,                
    showstringspaces=false,
    showtabs=false,                  
    tabsize=2
}
\title{Intern-Atlas: A Methodological Evolution Graph as Research Infrastructure for AI Scientists}

\author[1,2*]{Yujun Wu}
\author[1,3*]{Dongxu Zhang}
\author[1,4*]{Xinchen Li}
\author[1,5*]{Jinhang Xu}
\author[6]{Yiling Duan}
\author[1,7]{Yumou Liu}
\author[1,8]{Jiabao Pan}
\author[1]{Qiyuan Zhu}
\author[7]{Xuanhe Zhou}
\author[9]{Jingxuan Wei}
\author[4]{Siyuan Li}
\author[4]{Jintao Chen}
\author[1]{Conghui He}
\author[1,\dagger]{Cheng Tan}

\affiliation[1]{Shanghai Artificial Intelligence Laboratory}
\affiliation[2]{Peking University}
\affiliation[3]{Xi'an Jiaotong University}
\affiliation[4]{Zhejiang University}
\affiliation[5]{East China Normal University}
\affiliation[6]{Hunan University} 
\affiliation[7]{Shanghai Jiao Tong University}
\affiliation[8]{Shanghai University}
\affiliation[9]{University of Chinese Academy of Sciences}

\abstract{
Existing research infrastructure is fundamentally document-centric, providing citation links between papers but lacking explicit representations of methodological evolution. In particular, it does not capture the structured relationships that explain how and why research methods emerge, adapt, and build upon one another. With the rise of AI-driven research agents as a new class of consumers of scientific knowledge, this limitation becomes increasingly consequential, as such agents cannot reliably reconstruct method evolution topologies from unstructured text. We introduce Intern-Atlas, a methodological evolution graph that automatically identifies method-level entities, infers lineage relationships among methodologies, and captures the bottlenecks that drive transitions between successive innovations. Built from $1{,}030{,}314$ papers spanning AI conferences, journals, and arXiv preprints, the resulting graph comprises $9{,}410{,}201$ semantically typed edges, each grounded in verbatim source evidence, forming a queryable causal network of methodological development. To operationalize this structure, we further propose a self-guided temporal tree search algorithm for constructing evolution chains that trace the progression of methods over time. We evaluate the quality of the resulting graph against expert-curated ground-truth evolution chains and observe strong alignment. In addition, we demonstrate that Intern-Atlas enables downstream applications in idea evaluation and automated idea generation. We position methodological evolution graphs as a foundational data layer for the emerging automated scientific discovery.
}

\date{\today}
\correspondence{Cheng Tan, \email{tancheng@pjlab.org.cn}
\\
\\
  \parbox{\linewidth}{\centering
    \web~\href{https://intern-atlas.opendatalab.org.cn/}{\textbf{Website}} \quad
    \huggingface~\href{https://huggingface.co/datasets/OpenRaiser/Intern-Atlas}{\textbf{Dataset}}
  }
}

\makeatletter
\def\blfootnote{\xdef\@thefnmark{}\@footnotetext}
\makeatother

\begin{document}

\maketitle

\blfootnote{*Equal contribution.}

\section{Introduction}

\begin{figure*}[t]
\centering
\includegraphics[width=\textwidth]{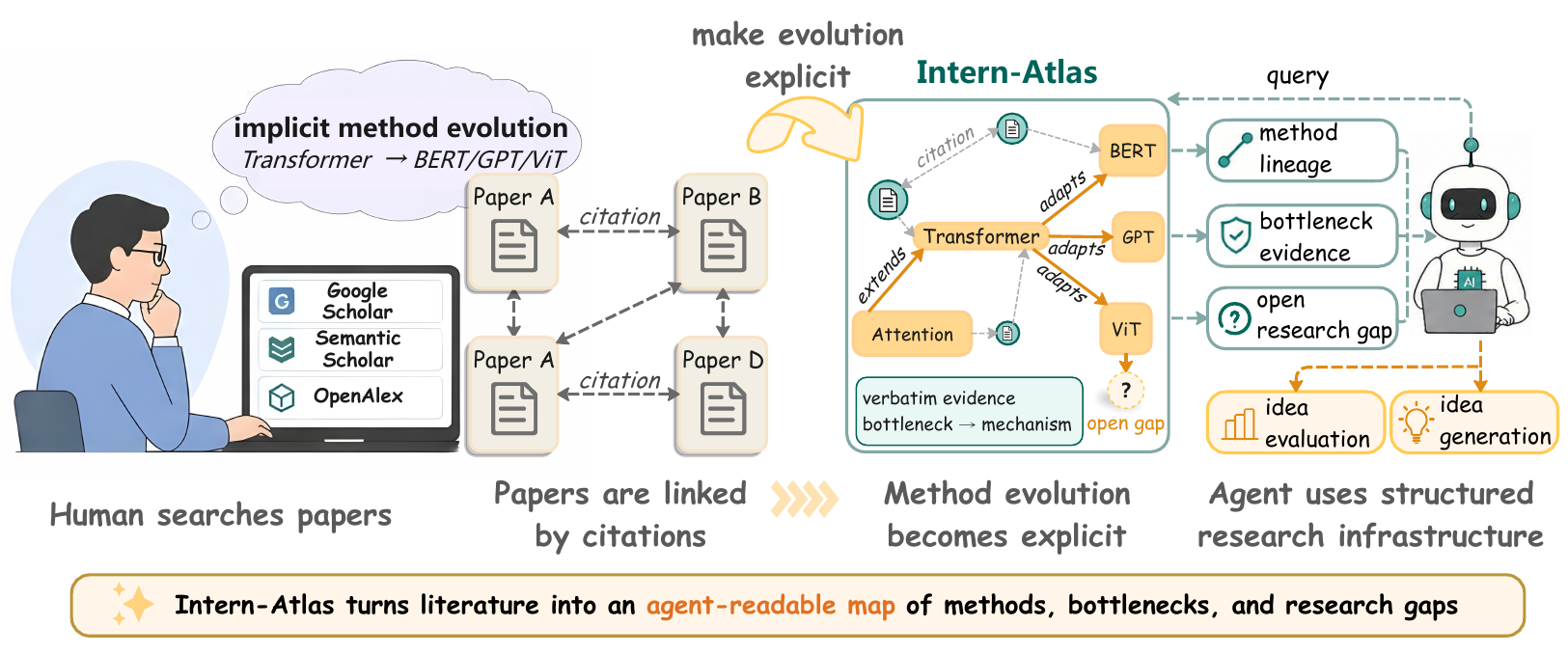} 
\caption{\textbf{From paper lists to an agent-readable methodology atlas.} Citation-based discovery tools (left) leave \emph{method} evolution (\textit{Transformer}\,$\to$\,\textit{BERT}/\textit{GPT}/\textit{ViT}) implicit in human expertise. \textsc{Intern-Atlas} (middle) makes it explicit as a typed graph with causal edge labels and verbatim bottleneck-to-mechanism evidence, providing an LLM agent (right) with direct queries for method lineage, bottleneck evidence, idea evaluation, and idea generation.}
\label{fig:intro}
\end{figure*}

Scientific progress in artificial intelligence has traditionally been documented through papers, and the infrastructure developed to organize this body of knowledge reflects a document-centric paradigm. Platforms such as Google Scholar~\cite{google_scholar}, Semantic Scholar~\cite{ammar2018construction}, and OpenAlex~\cite{priem2022openalex} treat the paper as the atomic unit: they index titles, abstracts, and citation counts, and they connect papers to one another through citation links. This paradigm has served human researchers for decades. To understand the emergence of vision transformers, for example, a researcher can retrieve a set of relevant papers from these systems, read them, and reconstruct the lineage from convolutional neural networks through self-attention mechanisms to modern architectures. The critical step in this workflow, extracting the structural relationships among methods from narrative text and assembling them into a coherent evolutionary picture, is performed entirely inside the researcher's head (Figure~\ref{fig:intro}, left).

A new class of knowledge consumer is now emerging. AI-driven research agents that automate the scientific pipeline from hypothesis generation to experimentation~\cite{lu2024ai,sanyal2025spark,fan2026deepinnovator} represent a fundamentally different kind of knowledge consumer. Unlike humans, these agents cannot reliably reconstruct method evolution topologies from unstructured text. Their parametric memory is a lossy compression that underrepresents low-frequency or long-tail methodological knowledge.Their autoregressive inference operates as fixed-depth forward computation rather than explicit graph traversal, limiting their ability to enumerate branching method spaces. Most critically, they lack the capacity to distinguish between genuine gaps in the research landscape and gaps in their own internal representations, since both manifest as an absence of relevant activations. As a result, their most persistent limitation lies in idea generation, where output quality depends on structural understanding of the methodological landscape: not only which methods exist, but how they evolved, what constraints they addressed, and which directions remain unexplored.

This gap reflects a recurring historical pattern. Structured knowledge infrastructure becomes essential not when humans require it, but when automated systems emerge that cannot operate effectively over unstructured data. The Protein Data Bank (PDB)~\cite{berman2000protein} standardized protein structures decades before AlphaFold~\cite{jumper2021highly} revealed their full value as machine-consumable training data. ImageNet~\cite{deng2009imagenet} organized visual data into hierarchical labels prior to the widespread adoption of convolutional neural networks that depended on large-scale structured annotations. In each case, the emergence of a new computational consumer transformed latent structure into an explicit requirement. A similar inflection point is now occurring in scientific methodology: \textit{AI research agents have emerged, yet the structured data layer required to support them remains absent}.

We introduce Intern-Atlas, a methodological evolution graph that fills this infrastructure gap. Atlas processes papers from top-tier AI venues, extracts method entities with alias resolution, classifies every citation edge into semantic types, and grounds each non-background edge in a verbatim quote with structured bottleneck and mechanism annotations. The resulting graph forms a queryable causal topology of methodological evolution, providing a persistent data layer that can be directly consumed by downstream systems (Figure~\ref{fig:intro}, right). Beyond graph construction, identifying meaningful evolution chains introduces an additional challenge. Methodological progress forms a directed acyclic graph: a single method such as the Transformer~\cite{vaswani2017attention} gives rise to multiple branches, including BERT~\cite{devlin2019bert}, GPT~\cite{radford2018improving,radford2019language,brown2020language}, Vision Transformer~\cite{dosovitskiy2020image,khan2022transformers,han2022survey}, and DETR~\cite{carion2020end}. Greedy traversal strategies that commit at each branching point discard alternative trajectories. To address this, we propose Self-Guided Temporal Monte Carlo Tree Search (SGT-MCTS), which balances exploitation of high-confidence paths with exploration of under-visited branches while enforcing temporal coherence, enabling more faithful reconstruction of methodological evolution.

We evaluate Intern-Atlas along three complementary dimensions. We first assess intrinsic graph quality by quantitatively comparing constructed evolution chains against expert-curated chains. We then examine its utility for idea evaluation, testing whether graph-grounded scoring signals exhibit monotonic alignment with human tiers. Finally, we investigate its value for idea generation by comparing outputs produced with and without access to graph-derived evolutionary context. Taken together, our results suggest that methodological evolution graphs can serve as a foundational data layer for the emerging ecosystem of automated scientific discovery.

\section{Related Work}

\subsection{Scientific Knowledge Graphs}
Tracing knowledge flow through citation networks has a long history. Main Path Analysis~\cite{hummon1989connectivity} identifies heavily traversed citation routes via flow-based traversal counts, representing the earliest formalization of evolution chain reconstruction. CiteSpace~\cite{chen2006citespace} detects research fronts through keyword burst analysis and co-citation clustering. Both operate on papers or keywords as atomic units, with edges defined by statistical co-occurrence rather than semantic causality.
Modern large-scale platforms have significantly expanded the metadata available for scientific literature. OpenAlex~\cite{priem2022openalex} indexes hundreds of millions of works; Semantic Scholar~\cite{ammar2018construction} and S2ORC~\cite{lo2020s2orc} augment standard citation graphs with contextual sentences and influential-citation markers; and Papers With Code~\cite{nanoresearch2026} introduces structured Task-Dataset-Metric triples. While these platforms serve as invaluable foundations, their network edges remain fundamentally untyped at the methodological level: a citation from paper A to paper B indicates general relevance but carries no machine-consumable structural information regarding whether A \textit{extends} B's architecture, \textit{resolves} B's specific bottleneck, or merely cites B as background context.
{Intern-Atlas} bridges this critical infrastructure gap by shifting the fundamental unit of analysis from monolithic documents to granular method entities. By coupling a robust methodology registry (including alias resolution) with explicitly typed causal edges (e.g., \textit{extends}, \textit{improves}, \textit{replaces}) grounded in verbatim textual evidence, Intern-Atlas transforms flat citation networks into a directed, queryable topology of methodological evolution. Ultimately, this structural shift provides the explicit, machine-consumable data layer required for automated AI research agents to systematically traverse and reason over the scientific landscape.

\subsection{Evaluating Research Ideas}
Evaluating research ideas is inherently noisy. Human evaluation exhibits high variance, demonstrated by low inter-annotator agreement~\cite{si2024can} and inconsistent peer reviews~\cite{beygelzimer2023has}. Automating this with Large Language Models (LLMs) introduces systematic bias: LLM-judged novelty correlates \textit{negatively} with scientific impact~\cite{latimer2025hindsight}, as models inherently favor safe, highly connected concepts~\cite{gu2024generation} due to their reliance on parametric familiarity rather than structural gap reasoning. 
Recent literature formalizes multi-dimensional evaluation criteria: AI Idea Bench~\cite{qiu2025ai} models novelty as historical difference and contemporary influence penalized by conformity; IdeaBench~\cite{guo2025ideabench} exposes the rarity of simultaneously high novelty and feasibility; and the Ideation-Execution Gap~\cite{si2025ideation} demonstrates quality collapse when highly novel ideas lack practical grounding. Despite their theoretical value, these frameworks remain purely \textit{descriptive}, lacking the infrastructure for deterministic computation.
{Intern-Atlas} operationalizes these descriptive frameworks into \textit{executable scoring functions}. We translate theoretical metrics into explicit topological signals: HindSight's dimensions are used to determine graph weightings, SciMuse's connectivity findings are mapped to calculable disconnection ratios, and IdeaBench's formulaic components are explicitly grounded in our computable topology.

\subsection{LLM-based Scientific Ideation}
End-to-end automated research systems have rapidly evolved from template-driven pipelines to sophisticated multi-agent loops. Early iterations like AI Scientist v1~\cite{lu2024ai} demonstrated fully automated workflows but relied heavily on hand-crafted templates. Subsequent architectures have introduced dynamic problem-solving: AI Scientist v2~\cite{yamada2025ai} utilizes agentic tree search to generate workshop-accepted papers; CycleResearcher~\cite{weng2024cycleresearcher} closes the research-review-revision loop to reach preprint-level quality; Dolphin~\cite{yuan2025dolphin} integrates exception-traceback-guided debugging; and AIGS~\cite{liu2024aigs} explicitly embeds falsification principles into the research process.
Parallel to execution, automated idea generation has seen significant focus. Chain of Ideas (CoI)~\cite{li2024chain} structures literature into linear chains for LLM-guided extrapolation, while systems like Nova~\cite{hu2024nova} and SciMON~\cite{wang2024scimon} optimize iterative search for novelty. However, both ideation and execution agents share a critical structural bottleneck: they construct their knowledge representations from scratch at task launch. CoI assembles chains transiently within prompts; AI Scientist hardcodes knowledge into templates; and SPARK~\cite{sanyal2025spark} re-retrieves unstructured text per query. 
This repeated, transient reinvention across independent systems constitutes explicit behavioral evidence of a missing infrastructure layer. {Intern-Atlas} fills this void by serving as their foundational prerequisite, providing the persistent, queryable, and method-granular knowledge base that current agents fundamentally lack.

\begin{figure*}[t]
\centering
\includegraphics[width=\textwidth]{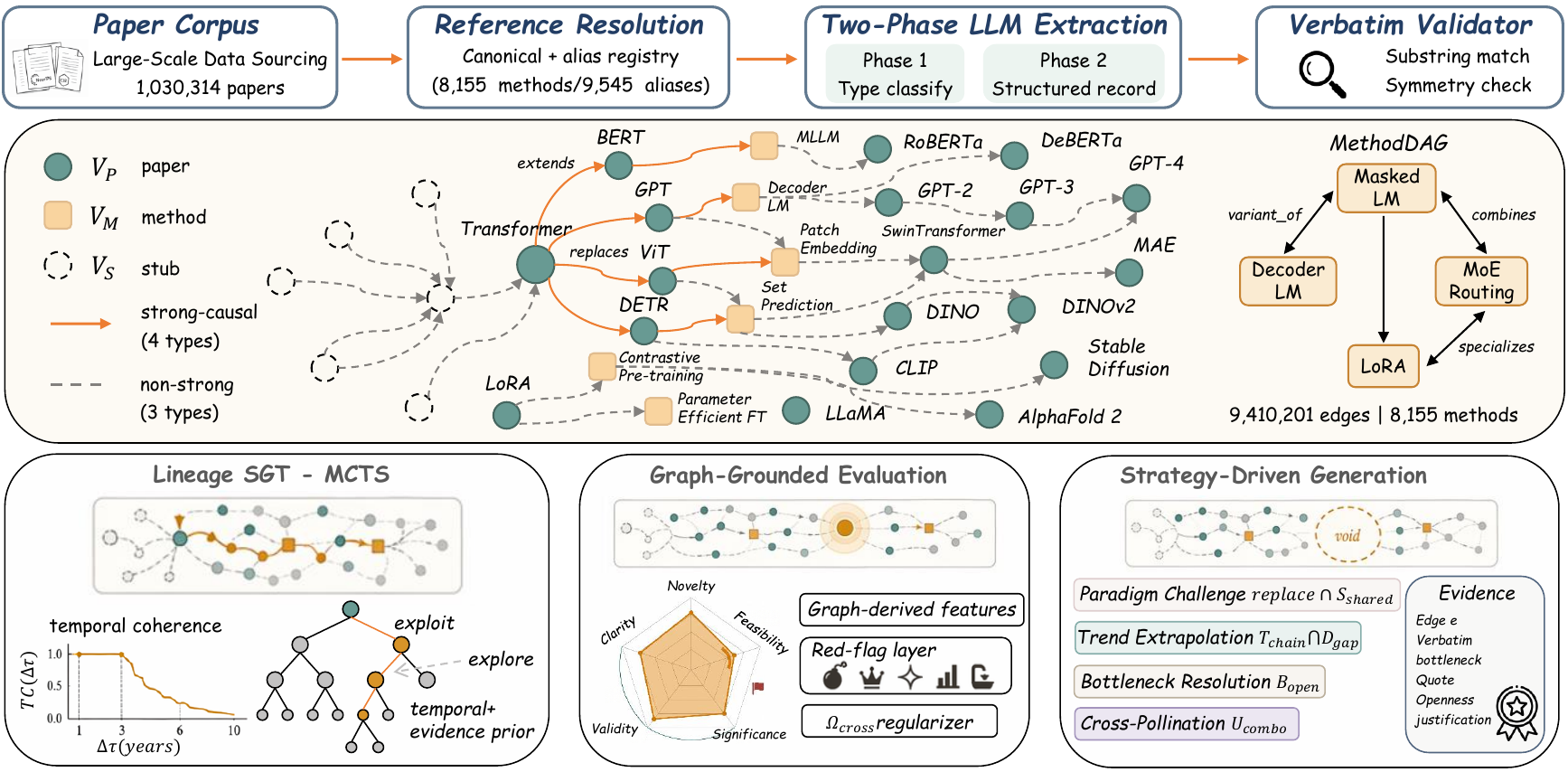} 
\caption{\textbf{Overview of \textsc{Intern-Atlas}.}
\textbf{Top:} From $1{,}030{,}314$ AI publications spanning conferences, journals, and arXiv preprints, we construct a typed methodology graph $G=(\mathcal{V},\mathcal{E},\tau,\rho)$ via reference resolution into papers, canonical methods, and stubs ($\mathcal{V}_P/\mathcal{V}_M/\mathcal{V}_S$; 8,155 methods, 9{,}545 aliases), two-phase LLM extraction (type classification $\to$ structured record), and a code-only verbatim validator. \textbf{Middle:} the resulting graph contains $9{,}410{,}201$ typed edges over 8,155 canonical methods; strong-causal edges (4 types, solid) induce the lineage subgraph $G_{\text{strong}}$, non-strong edges (3 types, dashed) supply retrieval context, and the projected method-level DAG $\mathcal{G}_M$. \textbf{Bottom:} \emph{Lineage SGT-MCTS} reconstruction with an evidence-physics prior over edge confidence and temporal coherence $TC(\Delta\tau)$; \emph{graph-grounded idea evaluation} on five axes with a parameter-free core function, red-flag detectors, and the cross-dimensional regularizer $\Omega_{\text{cross}}$; and \emph{strategy-driven idea generation} along four topological strategies, each proposal certified by a verbatim evidence record.}
\label{fig:pipeline}
\end{figure*}

\section{Method}
\label{sec:method}

\subsection{Overview}
\label{sec:overview}
Existing scholarly platforms~\cite{ammar2018construction, priem2022openalex} treat the paper as the atomic unit and connect papers through citations that carry no semantic information. For an AI research agent, this representation is impoverished: \textit{it cannot tell whether paper $A$ extends paper $B$'s architecture, addresses $B$'s specific bottleneck with a different mechanism, or merely mentions $B$ as background}. We introduce \textbf{Intern-Atlas}, a method-centric heterogeneous graph $G$ that closes this gap. Each edge in $G$ carries a type drawn from a seven-category vocabulary, and every causal edge additionally carries a direct quote from the citing paper that names the bottleneck the edge resolves.

The system (Figure~\ref{fig:pipeline}) operates in two stages. The first stage, \textbf{Methodological Graph Construction} (§\ref{sec:graph}), turns raw AI papers into $G$ by extracting method entities from text, typing every citation edge with one of the seven relations, and attaching to each causal edge a four-field evidence record: three text spans quoted directly from the citing paper (the bottleneck, the mechanism, and the trade-off), plus an LLM-reported confidence. The second stage, \textbf{Operators over the Graph} (§\ref{sec:operators}), instantiates three reasoning tasks on top of $G$: \textbf{lineage reconstruction} traces the causal ancestors and descendants of a target method through $G$; \textbf{idea evaluation} scores a research idea by where its constituent methods sit in $G$, treating the graph as a structural map of the field's methodological landscape; and \textbf{idea generation} searches $G$ for under-explored regions and proposes new ideas to occupy them.

\subsection{Methodological Graph Construction}
\label{sec:graph}

We build $G$ from a large-scale corpus of $1{,}030{,}314$ AI papers over the 1965--2025 period in three steps: (i) entity resolution turns raw text into typed nodes, (ii) edge typing classifies every citation into one of seven causal relations, and (iii) evidence extraction attaches a structured record to every causal edge.

\paragraph{Step 1: entity resolution.}
The graph has three node sets: papers $\mathcal{V}_P$ (the parsed corpus), methods $\mathcal{V}_M$ (one node per method), and stubs $\mathcal{V}_S$ (placeholders for cited works that fall outside the corpus, totaling $3{,}173{,}187$ nodes). $\mathcal{V}_M$ is seeded from a hand-curated list of well-known methods and expanded by an LLM proposer scanning the corpus for additional candidates. Because the same method appears under different names in raw text (e.g., Transformer vs. vanilla Transformer), we collect these names into a lookup table $\mathcal{A}: \text{string} \to \mathcal{V}_M$ that routes every observed name to the matching method node. The output of Step 1 is a citation graph in which every reference is mapped to a node in $\mathcal{V}_M \cup \mathcal{V}_S$.

\paragraph{Step 2: edge typing.}
Each resolved reference becomes a typed edge $e = (u, v)$ from a method node $u \in \mathcal{V}_M$ to its target $v \in \mathcal{V}_M \cup \mathcal{V}_S$. An LLM classifier reads the citing paper's text near the reference and assigns $e$ one of seven labels, ordered by decreasing causal strength: \emph{extends}, \emph{improves}, \emph{replaces}, \emph{adapts}, \emph{uses\_component}, \emph{compares}, and \emph{background}. We write $\text{type}(e)$ for this label. The first four categories denote a direct methodological lineage and form the \emph{strong-causal} subset $\mathcal{E}_{\text{sc}} \subseteq E$, on which lineage reconstruction operates; the remaining three are retained for retrieval context. Edges with $\text{type}(e) \neq \textit{background}$ are called \emph{causal edges} and proceed to Step 3.

\paragraph{Step 3: evidence extraction.}
For every causal edge, an LLM extractor populates a four-field evidence record:
\begin{equation}
\rho(e) \;=\; \bigl(\,b_e,\;m_e,\;t_e,\;c_e\,\bigr),
\label{eq:rho}
\end{equation}
where $b_e$ is the bottleneck addressed, $m_e$ is the mechanism, $t_e$ is the trade-off (each a text span quoted from the citing paper), and $c_e \in [0, 1]$ is an LLM-reported confidence. Each bottleneck is also tagged with one of 14 categories from a taxonomy $\mathcal{D}$ (e.g., \emph{computational cost}, \emph{long-range dependency}). This grouping lets the idea generator reason about which problem categories still lack a strong solution, rather than treating every bottleneck quote as unique. A deterministic post-checker discards any edge whose quoted spans fail substring match against the citing paper, whose endpoints violate publication-year ordering, or whose citing paper already carries an opposite-direction edge to the same target. Further details are in Appendix~\ref{app:graph}.

\subsection{Operators over the Methodological Graph}
\label{sec:operators}

All three operators begin the same way. Given a textual input $x$ (a target method, an idea description, or a research
 question), each operator first uses the lookup table $\mathcal{A}$ to find method nodes matching $x$, then expands
outward through BM25~\cite{10.1561/1500000019} keyword retrieval over $\mathcal{V}_P$ to collect related papers and
the typed edges among them. We denote the resulting localized context by:
\begin{equation}
\mathcal{C}_x \;=\; \mathrm{Retrieve}(x, G, \mathcal{A}).
\label{eq:retrieve}
\end{equation}
The three operators differ only in what they do with $\mathcal{C}_x$: lineage reconstruction (§\ref{sec:lineage})
traverses $\mathcal{E}_{\text{sc}}$ to recover evolution chains, idea evaluation (§\ref{sec:eval}) scores the input by
 where its constituent methods sit in $\mathcal{C}_x$, and idea generation (§\ref{sec:gen}) searches $\mathcal{C}_x$
for structural gaps and proposes ideas to fill them.

\subsubsection{Lineage Reconstruction with SGT-MCTS}
\label{sec:lineage}
Lineage reconstruction unfolds a method's evolutionary history. Starting from each seed in $\mathcal{S}(q) \subseteq \mathcal{C}_q$ that matches the query $q$ exactly, it returns a small set of evolution chains $\Pi_q$, where each $\pi \in \Pi_q$ is a directed path through the strong causal subgraph $(\mathcal{V}_M, \mathcal{E}_{\text{sc}})$ respecting publication-year order. Because hub methods have multiple successors, we adopt MCTS. We modify its UCT~\cite{kocsis2006bandit} selection rule to use the typed metadata in $G$. At node $u$, child $v$ is scored by:
\begin{equation}
\mathrm{SGT\text{-}UCT}(v) \;=\; \mathrm{UCT}(u, v) \;+\; \lambda \cdot \alpha_G(u, v),
\label{eq:sgt-uct}
\end{equation}
where $\mathrm{UCT}(u, v)$ is the standard exploit-explore criterion and $\alpha_G(u, v) = \mathrm{conf}(e_{u\to v})
\cdot \mathrm{TC}(\Delta\tau_{uv})$ is a graph-aware prior built from two factors. The first, $\mathrm{conf}(e_{u\to
v}) \in [0, 1]$, is the LLM-reported edge confidence stored in $\rho(e_{u\to v})$. The second,
$\mathrm{TC}(\Delta\tau_{uv})$, is a function of the publication-year gap $\Delta\tau_{uv}$ peaked at $\Delta\tau \in
[1, 3]$ years. Their product favors children that are both well-evidenced and temporally plausible. Backward and forward chains from each seed are joined and ranked by:
\begin{equation}
\mathrm{rank}(\pi) \;=\; w_\ell\,\tfrac{|\pi|}{L_{\max}} + w_c\,\overline{\mathrm{conf}}(\pi) +
w_m\,\overline{N}(\pi),
\label{eq:rank}
\end{equation}
combining three complementary terms. $|\pi|/L_{\max}$ is the chain's normalized length and rewards longer chains. $\overline{\mathrm{conf}}(\pi)$ is the mean edge confidence in $\pi$ and rewards chains whose every step is well-evidenced. $\overline{N}(\pi)$ is the mean MCTS visit count in $\pi$, following HopRAG~\cite{liu2025hoprag} in rewarding chains that many independent rollouts converge on. To recover parallel evolutions at hubs, the lineage operator re-runs from every branch point with covered edges masked, yielding
 the final $\Pi_q$.

\subsubsection{Graph-Grounded Idea Evaluation}
\label{sec:eval}
Idea evaluation places a research idea. Taking a textual description $d$ as input, it produces a final score $s^*(d, G)$ that reflects where $d$'s constituent methods sit on this landscape. Free-text LLM judges reward shallow stacking of popular method names; their novelty assessments correlate \emph{negatively} with eventual scientific impact while feasibility ratings correlate positively~\cite{latimer2025hindsight, gu2024generation}. Our evaluator therefore computes every per-dimension score through a deterministic graph query.

\paragraph{Per-dimension scoring.}
Each evaluation dimension probes a different structural property of $d$'s methods inside $G$, so its score can be computed directly from graph statistics rather than text-based LLM reasoning. The evaluator first resolves the methods named in $d$ into nodes $M_d \subseteq \mathcal{V}_M$ via the lookup table $\mathcal{A}$, then scores each of five dimensions (Novelty $N$, Feasibility $F$, Significance $S$, Validity $V$, Clarity $C$) independently from the position of $M_d$ in $d$'s retrieved context $\mathcal{C}_d$:
\begin{equation}
s_k(d, G) \;=\; \mathrm{clip}_{[1, 10]}\Bigl(b_k + \sum_{j} w^{(k)}_j \cdot \phi^{(k)}_j(M_d, \mathcal{C}_d)\Bigr),
\quad k \in \{N, F, S, V, C\}.
\label{eq:eval-raw}
\end{equation}
For each dimension $k$, the score combines a baseline $b_k$ with a sum of weighted graph statistics, where each
$\phi^{(k)}_j$ is a deterministic statistic computed from $M_d$ and $\mathcal{C}_d$, and $w^{(k)}_j$ is its fixed scalar weight. One of the summed terms $\phi^{(F)}_j$ is an \textit{anti-stacking} statistic that down-weights ideas dominated by overstudied methods, directly countering the LLM's positive feasibility bias toward popular method stacks. The complete catalogue of $\phi^{(k)}_j$ across the five dimensions is detailed in Appendix~\ref{app:eval-signals}.

\paragraph{Cross-dimensional aggregation.}
Reducing the five dimensions to a single verdict requires more than a weighted average: certain dimension combinations carry qualitative meaning that a linear aggregation cannot express. The evaluator then aggregates the five raw scores into a final score by combining a linear weighted sum with a non-linear correction:
\begin{equation}
s^*(d, G) \;=\; \mathrm{clip}_{[1, 10]}\bigl(\mathbf{w}^{\top}\mathbf{s} + \Omega_{\text{cross}}(\mathbf{s})\bigr),
\quad \mathbf{s} = (s_N, s_F, s_S, s_V, s_C).
\label{eq:eval-agg}
\end{equation}
Here $\mathbf{w}$ is a fixed weight vector and $\Omega_{\text{cross}}$ implements four hand-designed conjunctive penalties. A representative penalty fires when an idea scores high on Novelty but low on Feasibility, reflecting that novel but infeasible typically signals a flawed core proposal regardless of its Significance and Validity scores. Finally, an optional LLM reviewer may \textit{only lower} $s^*$, providing a one-sided veto that catches occasional errors in the graph-based scoring without reintroducing the LLM's upward bias.

\subsubsection{Strategy-Driven Idea Generation}
\label{sec:gen}

While evaluation asks ``how good is this idea?'', generation asks the inverse question: ``which new idea should be proposed?'' Operating on the retrieval context $\mathcal{C}_q$ ($q$ is a research method of interest), the generator scans for under-explored structural regions and returns a candidate research proposal $d'$ tied to a concrete bottleneck the literature has yet to solve. Grounding the search in graph structure, rather than free-text reasoning, prevents the LLM from fabricating proposals out of nothing.

\paragraph{Structural gap patterns.}
From $\mathcal{C}_q$ and lineage chains $\Pi_q$, the generator extracts four structural patterns that together pinpoint where the literature has unfilled research opportunities. These patterns (\textit{open axes}, \textit{recent improvement directions}, \textit{sacrifice axes}, and \textit{disconnected pairs}) are bundled into a structural summary $\Phi(q, G)$ via deterministic graph queries (no LLM call).

\paragraph{Strategy-conditioned proposal.}
To translate $\Phi$ into a concrete proposal, we pair each pattern type with a dedicated generation strategy: \textit{Bottleneck Resolution} for open axes, \textit{Trend Extrapolation} for recent improvement directions, \textit{Cross-Pollination} for disconnected pairs, and \textit{Paradigm Challenge} for sacrifice axes. This one-to-one mapping converts the LLM's task from open-ended ideation to constrained completion. The chosen strategy $\sigma$ fixes both the structural target and the proposal's high-level form, leaving the LLM only to fill in technical specifics:
\begin{equation}
d' \;=\; \mathrm{Generate}(\Phi(q, G), \sigma).
\label{eq:gen}
\end{equation}
The strategy alone, however, does not stop the LLM from inventing a bottleneck and dressing it up in the right form. We close this loophole by requiring every proposal to carry an evidence certificate: a triple $(e, b_e, j)$ where $e$ is a specific causal edge from $\mathcal{C}_q$, $b_e$ is the bottleneck text already stored in $\rho(e)$ (reproduced exactly), and $j$ explains why this bottleneck is still unresolved. Before a proposal is returned, we exact-match $b_e$ against the stored quote in $\rho(e)$; if it fails, the LLM output is discarded and a deterministic fallback emits a minimal valid proposal from graph content. The result is that every returned proposal is backed by a real edge in $G$ with verifiable bottleneck evidence: no fabricated motivations leak through, and the system always returns a valid proposal.


\section{Experiment}
\label{exp}

\begin{figure*}[t]
\centering
\includegraphics[width=\textwidth]{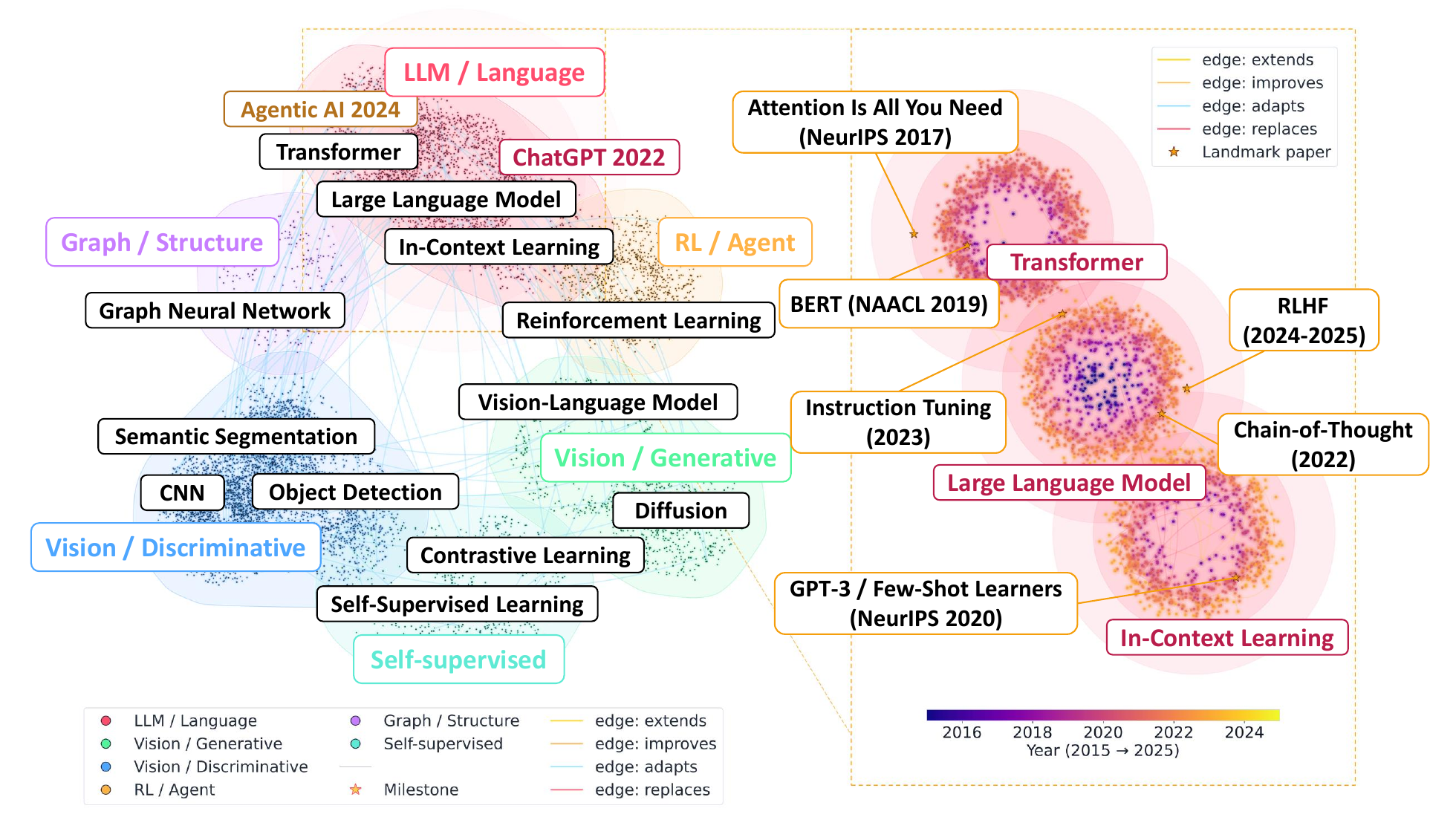}
\caption{\textbf{Intern-Atlas method landscape.}
\emph{Left:} AI papers cluster into six paradigm continents connected by typed evolution edges (extends/improves/adapts/replaces). For visual clarity, only a representative subset of nodes is displayed.
\emph{Right:} Detailed view of the LLM continent with landmarks such as Attention, BERT, GPT-3, CoT, Instruction Tuning, and RLHF.}
\label{fig:intern-atlas-overview-half}
\end{figure*}

Figure~\ref{fig:intern-atlas-overview-half} provides a high-level view of the Intern-Atlas method landscape, showing major paradigm clusters and their evolution links. We evaluate Intern-Atlas from two perspectives: whether the graph reliably represents methodological evolution, and whether this structure is useful for downstream research tasks. For graph quality, Sec.~\ref{exp01} compares the constructed graph with survey-derived expert references and tests whether SGT-MCTS can recover coherent method lineages from the graph. For downstream utility, Sec.~\ref{exp02} studies whether graph-derived evidence supports idea evaluation, while Sec.~\ref{exp03} studies whether retrieved method lineages help generate better research ideas. Together, these experiments evaluate Intern-Atlas at both representation and application level.

\subsection{Evaluating Graph Construction and Lineage Reconstruction}
\label{exp01}

This section evaluates whether Intern-Atlas captures meaningful method-evolution structure. The evaluation covers two aspects: whether the graph contains the key methods and evolution relations described in survey-derived benchmark, and whether SGT-MCTS can recover method lineages.

\paragraph{Method-evolution Benchmark and Metrics.} We build a benchmark from 30 high-impact survey papers covering major AI subfields, containing 30 survey-derived method-evolution graphs with 2,268 nodes, 1,462 edges, and 133 evolution chains. Each graph consists of method nodes and evolution relations from the corresponding survey, and the extracted chains serve as references for lineage reconstruction evaluation. For graph construction evaluation, we report Node Match Ratio (NMR), measuring how many survey methods can be matched to Intern-Atlas, and Edge Reachable Ratio (ERR), measuring how many method relations can be recovered as directed paths in Intern-Atlas. We further report Path Semantic Correctness (PSC), which measures whether the recovered path is semantically correct compared to the reference method-evolution relation. For lineage reconstruction evaluation, the chains retrieved by each method are compared with the reference chains using Node Recall (NR), Edge Recall (ER), and Chain Alignment Score (CAS). NR and ER measure the coverage of reference methods and reference transitions, while CAS measures whether the retrieved methods preserve the ordering of the reference chain. More details are provided in Appendix~\ref{app:survey-benchmark}.

\begin{figure}[t]
    \centering
    \begin{subfigure}{0.49\textwidth}
        \centering
        \includegraphics[width=\textwidth]{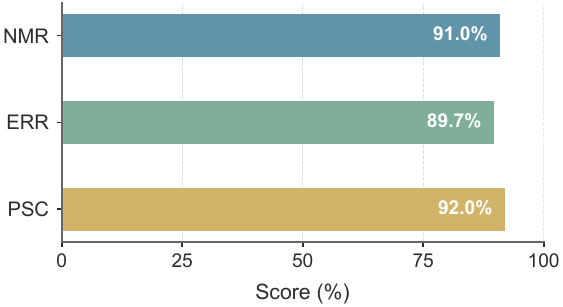}
        \caption{Graph quality scores}
        \label{fig:four_a}
    \end{subfigure}
    \hfill
    \begin{subfigure}{0.49\textwidth}
        \centering
        \includegraphics[width=\textwidth]{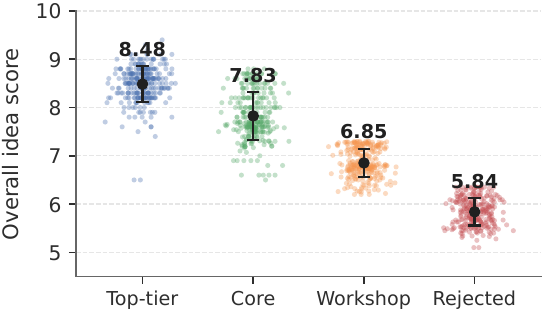}
        \caption{Overall tier scatter}
        \label{fig:four_d}
    \end{subfigure}

    \caption{\textbf{Graph construction quality and graph-grounded idea evaluation.}
  (a) Static graph quality on the method-evolution benchmark: node match ratio (NMR), edge reachable ratio (ERR), and path semantic correctness (PSC).
  (b) Distribution of Overall scores across strata (300 papers each); black markers indicate the sample mean $\pm$ 1 s.d.}
    \label{fig:main_four_results}
\end{figure}


\paragraph{Evaluation of SGT-MCTS algorithm.} 
\begin{wraptable}{r}{0.38\textwidth}
\centering
\vspace{-1.5em}
\caption{Lineage-search algorithm performance. `@$k$` denotes beam width or RW rollouts.}
\label{tab:lineage-search-results}
\begin{tabular}{lcccc}
\toprule
\textbf{Method} & \textbf{NR} & \textbf{ER} & \textbf{CAS} \\
\midrule
Beam@1 & 41.0 & 18.6 & 41.0 \\
Beam@5 & 43.4 & 21.6 & 43.4  \\
Beam@10 & 44.9 & 23.2 & 44.9 \\
RW@5 & 28.1 & 0.7 & 28.1 \\
\rowcolor{lightblue} SGT-MCTS & \textbf{84.8} & \textbf{79.0} & \textbf{84.8} \\
\bottomrule
\end{tabular}
\vspace{-1.0em}
\end{wraptable}
Table~\ref{tab:lineage-search-results} compares SGT-MCTS with two graph-search baselines. Beam search expands all valid next-hop methods at each step and keeps the top-$k$ highest-scored partial chains. Random walk starts from the same seed method and repeatedly samples a valid next-hop method uniformly at random until the depth limit or a dead end. Compared with the strongest baseline, Beam@10, it improves NR by 39.9 points, ER by 55.8 points, and CAS by 39.9 points, indicating better recovery of the reference methods, transitions, and ordering in target evolution chains. These results support the effectiveness of SGT-MCTS for method-lineage reconstruction. Given the same graph and seed method, SGT-MCTS finds chains that are much closer to the expected evolution lineage of the method than the baselines.

\paragraph{Analysis of graph construction quality.}
Figure~\ref{fig:four_a} shows Intern-Atlas performance on method-evolution benchmark across method coverage, relation reachability, and semantic correctness. The NMR of 91.0\% indicates that most methods in the reference graphs can be matched to nodes in Intern-Atlas.  The ERR of 89.7\% further indicates that most reference evolution relations can be recovered in Intern-Atlas as directed paths from the source method to the target method. These results suggest that Intern-Atlas captures both method entities and their evolution relations, rather than only isolated paper or method records. The PSC of 92.0\% further confirms that these recovered paths capture correct evolution semantics. The remaining mismatches are expected because the survey-derived graphs and Intern-Atlas are constructed with different procedures and levels of granularity. Overall, these results show that Intern-Atlas can correctly construct method-evolution structures.

\subsection{Evaluating Graph-Grounded Idea Evaluator}
\label{exp02}

\paragraph{Experimental details.}
We evaluate Intern-Atlas as an infrastructure for automated idea evaluation using the \textbf{Strata Dataset}, which contains 1,200 papers (300 per stratum) across four outcome categories: (1) Top-tier AI conferences (ICLR 2026, ICML 2025, NeurIPS 2025), (2) Core AI conferences (AAAI 2026, IJCAI 2025), (3) Workshop papers (from ICLR 2026), and (4) Rejected submissions (from ICLR 2026). For each paper, we extract a concise summary of its core idea as the evaluator input. Our evaluator (described in Sec.~\ref{sec:eval}) then assesses these extracted ideas, outputting scores across five dimensions (novelty, feasibility, significance, validity, and clarity) and an overall score. More details are in Appendix~\ref{app:idea-eval-data}.

\paragraph{Validation across publication strata.}
We use the Strata Dataset to examine whether Intern-Atlas scores follow the broad quality ordering of different publication strata. As shown in Table~\ref{tab:idea-eval-strata-transposed} and Figure~\ref{fig:four_d}, the overall score follows the expected order: top-tier conference papers receive the highest mean score (8.48), followed by core conference papers (7.83), workshop papers (6.85), and rejected submissions (5.84). This trend also holds for all five evaluation dimensions. The largest gaps appear in \textit{Significance} and \textit{Validity}, suggesting that the method-evolution graph is especially useful for judging whether an idea is tied to important and well-supported methodological directions. \textit{Novelty} also shows a clear decrease across strata, while \textit{Clarity} varies more mildly, which is reasonable because an extracted idea can be clearly stated even when its contribution is limited. Overall, these results suggest that Intern-Atlas can support idea-quality evaluation by using method-evolution structure as evidence, producing judgments that broadly align with academic consensus on publication quality.

\begin{table}[t]
\centering
\caption{Mean idea evaluation scores across publication strata. Scores are computed from extracted ideas. The overall score and all five dimensions decrease from top-tier conference papers to rejected submissions, with the largest gaps appearing in Significance and Validity.}
\label{tab:idea-eval-strata-transposed}
\begin{tabularx}{\linewidth}{l *{6}{>{\centering\arraybackslash}X}}
\toprule
& \textbf{Novelty} & \textbf{Feasibility} & \textbf{Significance} & \textbf{Validity} & \textbf{Clarity} & \textbf{Overall} \\
\midrule
Top-tier conferences & \textbf{7.27} & \textbf{7.09} & \textbf{9.49} & \textbf{8.62} & \textbf{6.57} & \textbf{8.48} \\
Core conferences & 6.54 & 6.80 & 7.16 & 6.47 & 6.52 & 7.83 \\
Workshop papers & 5.01 & 6.38 & 5.77 & 6.16 & 6.17 & 6.85 \\
Rejected submissions & 4.56 & 5.50 & 4.91 & 5.27 & 5.54 & 5.84 \\
\bottomrule
\end{tabularx}
\end{table}

\begin{figure*}[t]
    \centering
    
    \begin{subfigure}[t]{0.5\textwidth}
        \centering
        \includegraphics[width=\textwidth]{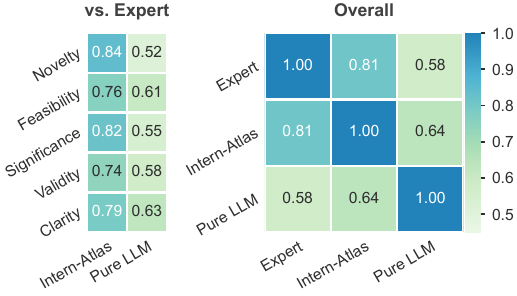}
        \caption{Human-alignment correlations}
        \label{fig:downstream_a}
    \end{subfigure}
    \hfill
    \begin{subfigure}[t]{0.49\textwidth}
        \centering
        \includegraphics[width=\textwidth]{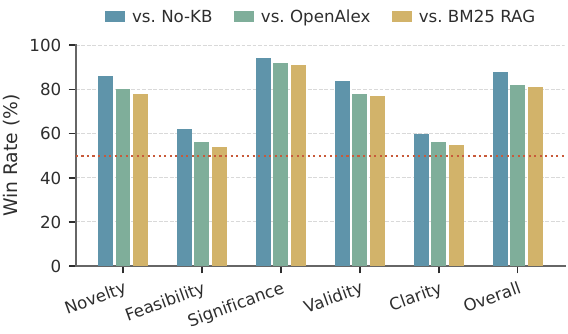}
        \caption{Pairwise generation win rate of Intern-Atlas}
        \label{fig:downstream_b}
    \end{subfigure}

    \caption{\textbf{Downstream utility of Intern-Atlas for idea evaluation and idea generation.}
  (a) Spearman correlations for idea evaluation: left, dimension-wise alignment of Intern-Atlas and a pure LLM judge with expert scores; right, overall-score correlations among expert judgments, peer reviews, Intern-Atlas, and the pure LLM baseline.
  (b) Pairwise win rates of Intern-Atlas against idea-generation baselines across five dimensions and overall quality.}
    \label{fig:downstream-results}
\end{figure*}

\paragraph{Human evaluation.}
For alignment with human expert review, we sample 100 idea profiles from the Strata Dataset and ask 10 AI PhD researchers to score them under the same five-dimensional rubric used by Intern-Atlas. We compare Intern-Atlas with a pure LLM-as-Judge baseline that uses only the idea profile and the scoring rubric as input. As shown in Figure~\ref{fig:downstream_a}, Intern-Atlas is more closely correlated with expert ratings than the pure LLM baseline across all five dimensions. The overall correlation is 0.81 for Intern-Atlas and 0.58 for the pure LLM baseline. The difference is most visible on \textit{Novelty} (0.84 vs. 0.52) and \textit{Significance} (0.82 vs. 0.55), where evaluating an idea requires comparing it with prior methods and understanding its position in methodological development. This suggests that large-scale method-evolution structure provides a more reliable basis for idea evaluation than prompting an LLM to judge from the idea text alone.

\subsection{Evaluating Graph-Grounded Idea Generator}
\label{exp03}

\begin{table}[t]
\centering
\caption{Idea-quality scores for generated proposals under different knowledge sources.}
\label{tab:exp03-idea-quality}
\begin{tabularx}{\linewidth}{l *{6}{>{\centering\arraybackslash}X}}
\toprule
\textbf{Method} & \textbf{Novelty} & \textbf{Feasibility} & \textbf{Significance} & \textbf{Validity} & \textbf{Clarity} & \textbf{Overall} \\
\midrule
No-KB & 4.85 & 6.52 & 5.70 & 4.00 & 7.21 & 5.78 \\
OpenAlex & 5.27 & 7.08 & 5.34 & 4.56 & 7.52 & 6.03 \\
Semantic Scholar & 5.40 & \textbf{7.32} & 5.39 & 4.70 & \textbf{7.75} & 6.18 \\
BM25 RAG & 5.39 & 7.30 & 5.39 & 4.66 & 7.70 & 6.15 \\
\rowcolor{lightblue} Intern-Atlas & \textbf{6.37} & 7.21 & \textbf{6.30} & \textbf{6.26} & 7.67 & \textbf{7.20} \\
\bottomrule
\end{tabularx}
\end{table}

\paragraph{Experimental details.}
To evaluate Intern-Atlas for idea generation, we employ the unified pipeline described in Sec.~\ref{sec:gen}, varying only the source of background knowledge. We compare Intern-Atlas against three baseline categories: (1) \textbf{No-KB}, which generates ideas directly from queries without external grounding; (2) \textbf{External Search}, using scholarly engines (OpenAlex and Semantic Scholar) as evidence sources; and (3) \textbf{Local RAG}, utilizing the same paper library as Intern-Atlas via standard BM25 retrieval. For input, we use 100 research questions curated by 10 AI PhD researchers. After generation, we score each generated idea using the idea-evaluation pipeline described in Sec.~\ref{sec:eval}. We further ask the same expert panel to compare Intern-Atlas against the baselines through blind pairwise win-rate evaluation. Additional details are provided in Appendix~\ref{app:idea-gen-queries}.

\paragraph{Evaluation Results.}
Table~\ref{tab:exp03-idea-quality} reports the idea-evaluation scores for ideas generated with different knowledge sources. Intern-Atlas achieves the best \textit{Overall} score (7.20) and ranks first in \textit{Novelty} (6.37), \textit{Significance} (6.30), and \textit{Validity} (6.26). The gain is most visible in \textit{Significance}, where method-evolution chains help the generator identify recurring limitations in prior work and propose ideas around more central research problems. The scores on \textit{Feasibility} and \textit{Clarity} are close across knowledge-grounded methods, since all methods use the same LLM generator and produce similarly readable proposals. Compared with the best baseline, Intern-Atlas improves the overall score by 1.02 points. Human-expert blind pairwise evaluation in Figure~\ref{fig:downstream_b} shows the same trend, with Intern-Atlas winning on overall quality by 88.0\%, 82.0\%, and 81.0\% against No-KB, OpenAlex, and BM25 RAG, respectively. Together, these results suggest that method-evolution context is more useful than document-level retrieval for generating high-quality research ideas.

\section{Conclusion}
We have introduced \textsc{Intern-Atlas}, a typed methodology graph that promotes flat paper citations into a queryable causal network of methodological evolution. Built from $1{,}030{,}314$ papers spanning AI conferences, journals, and arXiv preprints, the graph carries $9{,}410{,}201$ semantically typed edges with verbatim bottleneck-to-mechanism evidence on every causal relation, and exposes three operators over the same substrate: SGT-MCTS lineage reconstruction, graph-grounded idea evaluation, and strategy-driven idea generation. Empirically, Intern-Atlas recovers expert-curated evolution chains more faithfully than beam-search and random-walk baselines, produces quality signals that monotonically stratify across publication tiers and align with independent expert review, and generates ideas preferred over external scholarly search and standard RAG baselines under label-blind human judgment. Just as the Protein Data Bank preceded AlphaFold and ImageNet preceded modern visual recognition, we see methodological evolution graphs as a foundational data layer for the emerging ecosystem of automated scientific discovery. We release the graph and pipeline as open infrastructure, and hope future work will build on this direction toward AI research agents that reason over the full causal lineage of human knowledge rather than isolated papers.
\clearpage
\bibliography{ref}
\bibliographystyle{unsrt}

\clearpage
\appendix

\section*{Appendix Table of Contents}
\begingroup
\hypersetup{linkcolor=black}
\startcontents[appendix]
\printcontents[appendix]{}{1}{\setcounter{tocdepth}{2}}
\endgroup

\vspace{1em}
\clearpage
\section{Graph Construction}
\label{app:graph}

\subsection{Schema}

\paragraph{Node types.}
$\mathcal{V}$ partitions into three disjoint classes:
\begin{itemize}
\item Paper nodes $\mathcal{V}_P$: parsed full texts within the collection window, with resolved references.
\item Method entity nodes $\mathcal{V}_M$: canonical methods, each tagged with a canonical name; surface-form
mapping is handled by the alias registry $\mathcal{A}$ below.
\item Stub nodes $\mathcal{V}_S$: cited papers outside the collection window, retained as metadata-only placeholders
so that historical lineages remain reachable during SGT-MCTS.
\end{itemize}
The publication-year map $\tau:\mathcal{V}_P\cup\mathcal{V}_S\to\mathbb{Z}$ is populated from DOI / OpenAlex
metadata; missing entries are treated as a neutral default by the temporal-coherence function in Appendix~\ref{app:mcts}.

\paragraph{Bottleneck dimension taxonomy $\mathcal{D}$.}
The 14 axes in Table~\ref{tab:dim-taxonomy} were bootstrapped by clustering 500 randomly sampled bottleneck quotes
from a pilot run on NeurIPS-2024 full texts and consolidated by the authors into a minimal spanning set. Their
operational definitions are reproduced verbatim in the Phase-2 extraction prompt and are used wherever a record
field is typed by $\mathcal{D}$.


\begin{table}[h]
\centering
\caption{The 14-axis bottleneck dimension taxonomy $\mathcal{D}$. Each axis names a fundamental cost or quality of
methodological design; axes type the bottleneck and impact fields of $\rho(e)$ and feed the
feasibility/significance signals of the idea evaluator. Definitions are reproduced verbatim in the Phase-2
extraction prompt.}
\label{tab:dim-taxonomy}
\begin{tabularx}{\linewidth}{l X}
\toprule
Dimension & Operational definition \\
\midrule
computational complexity & asymptotic or wall-clock compute at fixed scale \\
memory efficiency & peak activation / parameter memory footprint \\
parallelization & degree of across-device or across-token parallelism \\
accuracy & task-level correctness or quality metric \\
generalization & out-of-distribution / cross-domain transfer \\
scalability & behavior as model / data / context size grows \\
data efficiency & sample complexity at fixed quality target \\
training stability & variance / divergence risk during optimization \\
inference speed & serving-time latency or throughput \\
expressiveness & function class or representational capacity \\
simplicity & implementation, conceptual, or interface simplicity \\
robustness & behavior under perturbation or adversarial input \\
hyperparameter sensitivity & outcome variance w.r.t.\ hyperparameter choice \\
training complexity & engineering difficulty of the training recipe \\
\bottomrule
\end{tabularx}
\end{table}

\paragraph{Edge vocabulary $\mathcal{T}$.}
Edges $\mathcal{E}\subseteq(\mathcal{V}_P\cup\mathcal{V}_S)^2$ carry a type from a nine-class causal vocabulary
(Table~\ref{tab:edge-types}), whose glossary is reproduced verbatim in the Phase-1 extraction prompt.

\begin{table}[h]
\centering
\caption{The seven causal edge types of $\mathcal{T}$. Definitions are reproduced verbatim in the Phase-1
extraction prompt; the first four form the strong-causal subset $\mathcal{T}_{\text{strong}}$ on which SGT-MCTS
performs lineage traversal.}
\label{tab:edge-types}
\begin{tabularx}{\linewidth}{l X}
\toprule
Edge type & Operational definition (Phase-1 prompt glossary) \\
\midrule
\texttt{extends} & adds a new capability or component on top of the cited method \\
\texttt{improves} & optimizes an existing component along $\ge 1$ dimension in $\mathcal{D}$ without altering the
core formulation \\
\texttt{replaces} & substitutes a load-bearing component with a qualitatively different mechanism \\
\texttt{adapts} & ports the cited method to a new domain, modality, or task setting \\
\texttt{uses\_component} & re-uses the cited method as an auxiliary module, not as the central contribution \\
\texttt{compares} & cites purely as a baseline or head-to-head comparison \\
\texttt{background} & cites as context, motivation, or non-methodological background \\
\bottomrule
\end{tabularx}
\end{table}

\paragraph{Strong-causal subgraph $G_{\text{strong}}$.}
$\mathcal{T}_{\text{strong}}=\{\texttt{extends},\texttt{improves},\texttt{replaces},\texttt{adapts}\}$ induces $G_{\text{strong}}=(\mathcal{V}_P\cup\mathcal{V}_S,\mathcal{E}_{\text{strong}},\tau,
\rho)$. Excluding \texttt{uses\_component} separates methodological evolution (inheriting the parent mechanism) from compositional reuse (borrowing only a module). Component reuse is preserved inside the
method-level DAG $G_M$ for evaluation and generation but does not drive lineage traversal.

\paragraph{Method-level DAG $G_M$.}
$G_M=(\mathcal{V}_M,\mathcal{E}_M)$ is derived from $\mathcal{E}$ by the deterministic projection in
Table~\ref{tab:projection}. Two further relations, \texttt{optimizes} and \texttt{inspired\_by}, are not produced
by citation-anchored extraction: \texttt{optimizes} typically surfaces in release-note or blog text lacking a
formal citation, while \texttt{inspired\_by} appears in narrative framing rather than bibliographic anchors. We
source both from hand-curated seed annotations, yielding $\mathcal{T}_M=\{\texttt{variant\_of}, \texttt{specializes}, \texttt{component\_of},\texttt{optimizes},\texttt{inspired\_by}\}$.

\begin{table}[h]
\centering
\caption{Deterministic projection from paper-level causal edges $\mathcal{T}$ to method-level relations
$\mathcal{T}_M$. \texttt{compares} and \texttt{background} are excluded as non-methodological, and
\texttt{uses\_component} is reversed in direction so that the cited method is encoded as a component of the citing
one.}
\label{tab:projection}
\begin{tabularx}{\linewidth}{l X}
\toprule
Causal type in $\mathcal{T}$ & Method-level type in $\mathcal{T}_M$ \\
\midrule
\texttt{extends}, \texttt{improves} & \texttt{variant\_of} \\
\texttt{adapts}, \texttt{replaces} & \texttt{specializes} \\
\texttt{uses\_component} & \texttt{component\_of} (direction reversed) \\
\texttt{compares}, \texttt{background} & excluded (non-methodological) \\
\bottomrule
\end{tabularx}
\end{table}

\paragraph{Alias registry $\mathcal{A}$.}
$\mathcal{A}:\mathcal{V}_M\to 2^{\Sigma^\ast}$ maps each canonical method to a set of surface forms. Lookup is
substring-based with case/punctuation normalization, word-boundary enforcement (preventing ``GPT'' matching inside
``lgpto''), longest-match priority (``GPT-4 Turbo'' over ``GPT-4'' over ``GPT''), version-suffix consolidation
(``-v2'', ``-Large'', etc.\ attach to the parent unless a distinct canonical node exists), and a manually curated
negative-surface list for ambiguous names (e.g., ``Mamba'' the state-space model versus the Python linter). The released registry contains $8,155$ canonical methods and $9,545$ aliases; construction details are in
\S\ref{app:corpus}.

\subsection{Corpus, PDF Pipeline, and Reference Resolution}
\label{app:corpus}

\paragraph{Collection window and venues.}
Intern-Atlas indexes a comprehensive corpus of major AI venues, journals, and preprint servers over the window 1965--2025, spanning core ML (NeurIPS, ICML, ICLR), computer vision (CVPR, ICCV, ECCV), NLP (ACL, EMNLP, NAACL), general AI (AAAI, IJCAI), and data/information (KDD, SIGIR). ICCV and ECCV run in alternating years, explaining their narrower temporal footprint; all other venues contribute three editions each. Main-conference, Findings (for ACL/EMNLP/NAACL), and accepted-workshop tracks are included uniformly.


\paragraph{PDF parsing pipeline.}
Full texts are obtained from the official proceedings (where openly licensed) and from OpenReview / arXiv
preprints as fallback. Each PDF is parsed by \textsc{Nougat-v1.0} for born-digital PDFs, with
\textsc{GROBID-v0.8.0} as a fallback for scanned or degraded files that Nougat fails on (3.7\% of the corpus). The
parser emits structured JSON with section headings, paragraph-level text, and a reference list. Downstream
extraction consumes only the \texttt{Introduction}, \texttt{Method}/\texttt{Approach}, and \texttt{Related Work}
sections of the citing paper; for cited references only the title and abstract are consumed in Phase 1, and the
title plus parsed bibliographic metadata in Phase 2.

\paragraph{Method entity curation.}
$\mathcal{V}_M$ is populated in two stages. First, a seed list of $247$ widely-used methods is hand-curated by
combining the Papers With Code method ontology with mentions enumerated in $18$ pilot review articles spanning
2020--2024. Second, an LLM expansion pass scans full-text Method sections for proper-noun method names following
capitalization and acronym heuristics; candidates are retained iff they co-occur with $\ge 3$ distinct papers'
Method sections and pass adjudication by one of the authors.

\subsection{Two-Phase Extraction Protocol}
\label{app:extraction}

Extraction decomposes into a high-recall Phase-1 classification over $\mathcal{T}$ followed by a Phase-2
record-completion pass over non-background candidates. We use \textsc{Qwen3.6-35B-A3B} as the production
extraction model for both phases and \textsc{Claude-Sonnect-4.6} as the audit model used only in the reliability
benchmarks. All calls use temperature $0$ and deterministic decoding; full prompts are released with the code
repository.

\paragraph{Phase 1 (classification).}
The model receives the citing paper's Introduction and Method sections, a list of references with title and
abstract, and the verbatim glossary of Table~\ref{tab:edge-types}. For each reference, it returns a JSON record
\texttt{\{type, evidence\_snippet\}}, with \texttt{background} reserved for citations that describe no
methodological relation. Phase 1 is tuned for high recall: the prompt explicitly instructs the model to prefer a
causal type over \texttt{background} in borderline cases, and the downstream validator together with Phase 2
suppresses false positives. Each prompt batches all references from one citing paper (mean $21.2$, max $96$); mean
input/output lengths are $5.8$k / $2.1$k tokens.

\paragraph{Phase 2 (record completion).}
Phase 2 consumes only the non-background candidates from Phase 1, batched 10 at a time together with the citing paper's Method section and the verbatim taxonomy of Table~\ref{tab:dim-taxonomy}. For each (citing, cited) pair already typed, the model fills in the structured record $\rho(e)=$\texttt{\{bottleneck:\{description, severity, verbatim\_quote, dimension\}, mechanism:\{description, type, verbatim\_quote\}, impact:\{improvement\_dim, sacrifice\_dim, tradeoff\_sentence\}, confidence\}}, with all \texttt{verbatim\_quote} fields required to be copied verbatim from the citing paper's text. Malformed or schema-violating batches are retried once with a repair prompt; a second failure drops the batch and logs it for manual inspection. Mean input/output lengths are $7.9$k / $2.8$k tokens.

\section{Lineage Reconstruction (SGT-MCTS)}
\label{app:mcts}

\paragraph{Hyperparameters.}
$c=\sqrt{2}$ (UCT exploration), $\lambda=0.5$ (prior weight), per-direction-per-seed budget $B=200$, rollout depth
cap $d_{\max}=5$, top-$K$ primary lineages $K=5$, deduplication node-Jaccard threshold $\theta_{\text{dup}}=0.8$,
dead-end backpropagation penalty $\eta=-0.05$.

\paragraph{Temporal coherence.}
\begin{equation}
\mathrm{TC}(\Delta\tau)=
\begin{cases}
0.40 & -1\le\Delta\tau<0 \\
0.85 & \Delta\tau=0 \\
1.00 & 1\le\Delta\tau\le3 \\
0.80 & 4\le\Delta\tau\le6 \\
\max(0.30,\,1.00-0.08(\Delta\tau-6)) & \Delta\tau>6 \\
0.70 & \tau\ \text{missing}
\end{cases}
\label{eq:tc}
\end{equation}
Edges with $\Delta\tau<-1$ in the search direction are hard-filtered prior to evaluation.


\paragraph{Algorithm.}
For each seed $s\in\mathcal{S}(q)$ and each direction in $\{\text{back},\text{fwd}\}$, MCTS runs for $B$
iterations on $G_{\text{strong}}$. Each iteration follows four standard phases: (i) selection by SGT-UCT
(Eq.~\ref{eq:sgt-uct}); (ii) expansion of the highest-confidence untried child after discarding cycles and edges
violating the hard temporal filter; (iii) greedy rollout that at each step picks the child maximizing
$\mathrm{conf}\cdot\mathrm{TC}$ and terminates at $d_{\max}$, a leaf, or a cycle, scored by $R(\pi)$;
(iv) backpropagation along the selection path, with an additional penalty $\eta$ to ancestors of dead-end leaves.
The two trees per seed are spliced through $s$ by concatenating each top-5 backward path (ranked by accumulated
$Q$) with each top-5 forward path; the resulting candidates are deduplicated under $\theta_{\text{dup}}$.


\paragraph{Branch discovery.}
After the top-$K$ primary lineages are finalized, SGT-MCTS is re-run from each \emph{branch point}, defined as a
node with $\ge 2$ strong-causal children in the search direction of which only one was traversed by the primary
set. Each re-run masks already-covered edges, uses budget $B/2$, and contributes its surviving lineages to the
output pool under the same deduplication rule, surfacing parallel evolutionary trajectories that pure greedy
traversal would collapse.

\section{Idea Evaluation}
\label{app:eval}

\subsection{Common Retrieval Routine}
\label{app:retrieval}

\paragraph{Context retrieval.}
Given $x$ (an idea $d$ or a query $q$), the routine returns the tuple $(M_x, P_x, E_x, B_x)$:
\begin{enumerate}
\item $M_x \subseteq \mathcal{V}_M$: canonical methods resolved from $x$ via the alias registry $\mathcal{A}$.
\item $P_x \subseteq \mathcal{V}_P$: the top-$K$ papers ($K=500$) under a hybrid score combining alias-count over
$M_x$ with BM25 keyword relevance on the remaining content words.
\item $E_x \subseteq \mathcal{E}$: non-\texttt{background} causal edges incident on $P_x$.
\item $B_x$: bottleneck records aggregated from $\rho(E_x)$.
\end{enumerate}

\subsection{Duplicate-Risk Detection}

A separate retrieval stack, invoked only by the Novelty signal, ranks $\mathcal{V}_P$ against $d$ in three stages.
Stage~1 (\emph{candidate pooling}) fuses a dense ranking from \textsc{BGE-small-en-v1.5} (384-dim cosine
similarity) with a sparse BM25-Okapi ranking ($k_1=1.5$, $b=0.75$) via Reciprocal Rank Fusion
($k_{\text{RRF}}=60$), yielding the top-20 candidates. Stage~2 (\emph{rerank}) re-scores those candidates with the
\textsc{ms-marco-MiniLM-L-12-v2} cross-encoder on $(d, p)$ pairs. Stage~3 (\emph{final score}) returns
$\mathrm{fused}(d,p)=0.5\,\mathrm{dense}(d,p)+0.5\,\sigma\!\big(\mathrm{rerank\_logit}(d,p)\big)$; the sparse
signal contributes only through Stage~1 pooling and does not enter the final fusion. Writing
$s=\max_p\mathrm{fused}(d,p)$, the Novelty signal applies a step penalty of $-0.5$ at $s\!\ge\!0.55$, $-1.5$ at
$s\!\ge\!0.65$, $-2.5$ at $s\!\ge\!0.75$, and $-4.0$ at $s\!\ge\!0.85$ (zero otherwise).

\subsection{Per-Dimension Signal Specifications}
\label{app:eval-signals}

\paragraph{Bases and weights.}
All five dimensions share base $b_k=5.0$. Aggregation weights are
$\mathbf{w}=(w_N,w_F,w_S,w_V,w_C)=(0.20,0.20,0.25,0.20,0.15)$. Signal contributions are additive on $b_k$; the sum
is clipped to $[1,10]$.

\paragraph{Per-dimension signals.}
Load-bearing components are listed below.

\emph{Novelty.} (i) Topological disconnection of $M_d$ in the method-co-utilization graph, up to $+2.0$;
(ii) mechanism-level Jaccard distance against $E_d$, $+[0,1.5]$; (iii) frontier-leaf bonus, $+0.8$;
(iv) duplicate-risk penalty $-\mathrm{pen}(d)\in[-4.0,0]$ from \S\ref{app:retrieval}.

\emph{Feasibility.} A sweet-spot maturity curve over each method's paper-count $\mathrm{pc}(m)$,
\begin{equation*}
\phi_F(m)=
\begin{cases}
1.5+1.5\cdot\tfrac{\mathrm{pc}(m)}{500} & \mathrm{pc}(m)\le500 \\
3.0-1.0\cdot\tfrac{\mathrm{pc}(m)-500}{1500} & 500<\mathrm{pc}(m)\le 2000 \\
1.5 & \mathrm{pc}(m)>2000
\end{cases}
\end{equation*}
averaged over $M_d$ and capped at $+3.5$. The $500/2000$ thresholds are the 60th and 90th percentiles of
$\mathrm{pc}(\cdot)$ on the released corpus; the non-monotonic shape blocks mashups from inflating $F$ via
pervasive terms. Auxiliary signals: full-text resource availability and a complexity penalty for $|M_d|\ge 4$.

\emph{Significance.} Time-decayed in-degree on $P_d$ (5-year half-life, non-background); frontier presence
($\ge 3$ non-background out-edges since 2021); and the \emph{method-frontier regularizer}, the dominant
anti-mashup term, with mean popularity over $M_d[:5]$ contributing $+2.5$ when $<300$ (niche) and decaying
linearly to $-2.0$ at $>1000$ (saturated).

\emph{Validity.} Bottleneck grounding against $B_d$ (matched dimension tag plus a description bigram, up to $+3.5$); ancestry consistency in $G_M$ within depth $4$; type-weighted edge density on $E_d$.

\emph{Clarity.} Method recognition rate; specificity peak at $|M_d|\in\{2,3\}$ decaying for $|M_d|\ge 6$; structural completeness across \{problem, approach, target\}; length adequacy on $20$--$200$ words.

\subsection{Cross-Dimensional Regularizer $\Omega_{\text{cross}}$}
\label{app:cross-dim}

\begin{table}[h]
\centering
\caption{Four empirical cross-dimensional priors comprising $\Omega_{\text{cross}}$. Each row triggers an additive
adjustment to the overall score $s^\ast$ when its conjunctive condition over the post-red-flag scores
$\mathbf{s}'$ holds, encoding patterns where one dimension's score becomes more or less informative in
conjunction with another.}
\label{tab:cross-dim}
\begin{tabularx}{\linewidth}{l X c}
\toprule
Prior & Trigger & Adjustment \\
\midrule
Ideation--execution gap & $s'_N\ge 7\ \land\ s'_F<4$ & $-0.6$ \\
Validity--feasibility coherence & $s'_V\ge 7\ \land\ s'_F\ge 7$ & $+0.2$ \\
Significance amplification & $s'_S\ge 6$ & $+0.4$ \\
& $5\le s'_S<6$ & $+0.2$ \\
Balanced coherence & $\max(\mathbf{s}')-\min(\mathbf{s}')\le 2\ \land\ \min(\mathbf{s}')\ge 5$ & $+0.3$ \\
\bottomrule
\end{tabularx}
\end{table}

\subsection{Graceful Fallback and Optional Adjudication}
\label{app:adjudication}

\paragraph{Graceful fallback.}
If $M_d=\emptyset$, the pipeline returns a population prior $s^\ast=6.5$, abstaining from the graph-grounded
functional rather than systematically penalizing purely theoretical contributions whose novelty is orthogonal to
the method registry. On a 147-idea pilot from NeurIPS-2024 submissions, text-only signals under $M_d=\emptyset$
gave Spearman $\rho=0.08$ ($p=0.31$) against peer-review scores, justifying abstention.

\paragraph{Adjudicator (optional).}
When enabled, two LLM calls modify $s^\ast$ via bounds rather than recompute it. \emph{Part A} (duplicate
verification): the adjudicator classifies the relationship between $d$ and the top-3 retrieved candidates as
\{duplicate, related, unrelated\}; the strongest verdict determines the restoration rate of the automatic
duplicate-risk penalty: $0\%$ for duplicate (full penalty retained), $60\%$ for related, $90\%$ for unrelated.
\emph{Part B} (per-dimension upper bound): the adjudicator rates the idea on coherence, novelty validity, and
claim plausibility on $1$--$10$; each rating acts as an upper bound on its mapped Atlas dimension via
$s_k\le\mathrm{llm\_score}_k+1.0$, with coherence $\to\{V,C\}$, novelty $\to N$, plausibility $\to\{F,S\}$. Any
sub-score below $3$ triggers a hard cap $s^\ast\le 6.0$. The adjudicator therefore never computes $s^\ast$
\emph{de novo}, preserving the zero-trainable-parameter property of the core scorer.

\section{Evaluation Setup and Benchmarks}
\label{app:setup}

\subsection{Method-evaluation Benchmark and Metrics}
\label{app:survey-benchmark}

\paragraph{Purpose and selection.}
We construct the method-evolution benchmark from 30 high-impact survey papers, using surveys as proxies for expert consensus on methodological evolution.
Candidate surveys are selected according to five criteria: authority, impact, classic status, subfield diversity, and method-centricity. In practice, we
prioritize surveys published in recognized venues, widely cited by the community, commonly used as field-level taxonomies, covering diverse AI subfields,
and containing explicit discussions of method evolution rather than only application summaries. These surveys provide external references for both static
graph coverage and lineage reconstruction evaluation.

\paragraph{Construction.}
Each survey is processed in two stages. First, an LLM extracts discussed methods, directed evolution relations, and narration-based method chains from the
survey text and taxonomy figures. Second, domain researchers manually audit the extracted results by removing hallucinated methods, correcting relation
directions, merging duplicate method names, and checking each retained relation against the original survey text or figure. The final benchmark contains 30
survey-derived method-evolution graphs with 2,268 nodes, 1,462 directed evolution edges, and 133 reference evolution chains.

\paragraph{Evaluation protocols.}
We evaluate the benchmark in two parts, following the main text.

\textbf{1. Static Graph Coverage.}
Given a survey-derived graph $G^\ast$ and Intern-Atlas graph $G$, we report Node Match Ratio (NMR), Edge Reachable Ratio (ERR), and Path Semantic
Correctness (PSC). NMR measures the fraction of survey methods matched to Intern-Atlas nodes. ERR measures the fraction of reference edges that can be
recovered as directed paths in $G$. PSC measures whether reachable paths preserve the intended method-evolution semantics, as judged against the survey
reference. Shortest-hop statistics are used only as a diagnostic for path compactness.

\textbf{2. Lineage Reconstruction.}
For each reference chain, every search method starts from the same seed and returns a candidate chain. We report Node Recall (NR), Edge Recall (ER), and
Chain Alignment Score (CAS). NR measures how many reference methods are recovered. ER measures how many adjacent reference transitions are recovered. CAS
measures how well the retrieved chain preserves the ordering of the reference chain.

\subsection{Lineage-Search Baselines}
\label{app:search-baselines}

To evaluate the lineage reconstruction capability of SGT-MCTS, we compare it against two standard graph-search baselines. All algorithms operate on the identical knowledge graph, originate from the same query seeds, and are constrained by the same maximum depth and computational budget. 

\begin{itemize}
\item \textbf{Beam Search:} Expands successors and selectively retains the top-$k$ partial paths at each step. It utilizes the identical local heuristic signals as SGT-MCTS but lacks global rollout-based exploration. This serves as a baseline for greedy semantic search, demonstrating the limitations of purely exploitative pathfinding without long-term planning. We evaluate beam widths of $k \in \{1, 5, 10\}$.
\item \textbf{Random Walk (RW):} Uniformly samples valid outgoing edges from the current node until the depth cap is reached. This process is repeated iteratively to match the total node-evaluation budget of SGT-MCTS. As a naïve exploration baseline, it isolates and validates the necessity of heuristic guidance and value estimation in our method.
\end{itemize}

All retrieved lineages from these baselines are evaluated against the same reference chains and metric suite described previously.

\subsection{Dataset and Baselines Details for Evaluating Idea Evaluator}
\label{app:idea-eval-data}
To validate the graph-grounded idea evaluator, we construct the Strata Dataset and apply a unified extraction protocol that standardizes the input format across all compared methods. We use the full dataset for publication-strata analysis and a 100-sample expert-rated subset for human-alignment evaluation.

\paragraph{The Strata Dataset.}
This dataset tests whether the evaluator's outputs follow the broad quality ordering reflected by different publication strata. We collected 1,200 papers evenly split across four tiers (300 each): (1) \textbf{Top-tier AI conferences}: accepted papers from ICLR 2026, ICML 2025, and NeurIPS 2025; (2) \textbf{Core AI conferences}: papers from AAAI 2026 and IJCAI 2025; (3) \textbf{Workshop papers}: submissions accepted to ICLR 2026 workshops; and (4) \textbf{Rejected submissions}: papers declined at ICLR 2026. All texts and metadata were obtained from OpenReview and publicly available proceedings. Venue and outcome labels are used only for aggregate analysis and are not provided to the evaluator.

\paragraph{Human-Rated Subset.}
For human-alignment evaluation, we sample 100 idea profiles from the Strata Dataset and ask 10 AI PhD researchers to rate each profile on a 1-10 scale across the same five dimensions used by Intern-Atlas: novelty, feasibility, significance, validity, and clarity. The experts are active AI researchers and are shown only the extracted idea profiles, not the paper source, venue, acceptance outcome, Intern-Atlas scores, or LLM baseline scores. We average expert ratings for each profile and compute Spearman correlations between the expert scores and the scores produced by Intern-Atlas and the pure LLM-as-Judge baseline.

\paragraph{Idea Extraction Protocol.}
Before scoring, we extract a standardized summary from each paper to ensure the evaluator focuses on the research idea itself rather than writing style. Specifically, given a paper's abstract, introduction, and core method sections, an LLM-based extractor produces a structured \emph{Idea Profile} with four fields: (1) \textbf{Problem Formulation}: the specific limitation or bottleneck the paper addresses; (2) \textbf{Core Innovation}: the key idea or mechanism proposed; (3) \textbf{Technical Implementation}: the main algorithmic components needed to realize the idea; and (4) \textbf{Target Setting}: the task or application domain.
The extractor is prompted to use a neutral tone and omit background exposition, rhetorical framing, and unsubstantiated claims, retaining only the factual methodological content. The evaluator then operates on this extracted profile rather than the raw paper text, which keeps the input format consistent across all compared methods and avoids confounding the evaluation with writing quality. The exact prompt is provided in the code release.

\paragraph{Pure LLM-as-Judge Baseline.}
This baseline employs Qwen3.5-397B-A17B to evaluate the standardized \emph{Idea Profile} in a zero-shot setting. It is prompted with the exact same five-dimensional 1--10 scoring rubric used by Intern-Atlas and the human experts, but relies strictly on its internal parametric knowledge. It computes the evaluation without access to external document retrieval engines or any structured method-evolution graph context.

\subsection{Experimental Details for Evaluating Idea Generation}
\label{app:idea-gen-queries}

\paragraph{Query Construction and Distribution.}
To standardize the evaluation of automated idea generation, we curated a diverse set of 100 high-quality research queries. To ensure clear constraints and evaluation objectives, each query follows a strict tripartite structure:
\begin{itemize}
\item \textbf{Domain Setting:} Establishes the specific AI subfield (e.g., \emph{``In retrieval-augmented generation...''}).
\item \textbf{Core Challenge:} Identifies the fundamental bottleneck or limitation of existing methods (e.g., \emph{``...multi-step retrieval is fundamentally restricted by high resource cost...''}).
\item \textbf{Expectation:} Specifies the desired capabilities of the target solution (e.g., \emph{``Therefore, there is a critical need to build a resource-efficient mechanism...''}).
\end{itemize}
To comprehensively evaluate the generative generalizability across different topics, these 100 queries are distributed across six major AI fields: NLP \& LLMs (34\%), General ML \& Optimization (19\%), Computer Vision (14\%), Reinforcement Learning \& Agents (14\%), Multimodal \& Speech (13\%), and Graph Learning (6\%). The full list of queries is provided in the supplementary material.

\paragraph{Human Expert Win-Rate Evaluation.}
To provide a rigorous, unbiased assessment of idea quality and compute the win rates, we assembled a panel of human experts, comprising active researchers with publication records in top-tier AI venues. The evaluation followed a strict double-blind pairwise comparison protocol. For each query, the experts were presented with the reference query alongside anonymized idea proposals generated by Intern-Atlas and the baselines. They were tasked with selecting the superior idea (or declaring a tie) based on overall methodological quality, specifically factoring in \emph{Novelty}, \emph{Feasibility}, and \emph{Significance}. Crucially, experts were permitted to consult external academic search tools during the evaluation to verify factual correctness and the true novelty of the proposed mechanisms.

\section{Case Studies}
\label{app:case-studies}

\subsection{Case Study on Lineage Search}
\label{app:case-study-search-resnet}

We use a familiar ConvNet lineage to illustrate the behavior of different search algorithms. All searches are seeded at \emph{ConvNeXt V2}. The reference
chain is a survey-level summary of the residual ConvNet backbone: \emph{VGG} scales ConvNets through deeper stacks of small convolutions; \emph{ResNet}
addresses the optimization difficulty of very deep networks with residual mappings; \emph{ResNeXt} extends residual blocks with aggregated transformations;
\emph{ConvNeXt} modernizes ResNet-style ConvNets using Transformer-era design choices; and \emph{ConvNeXt V2} further extends ConvNeXt with a fully
convolutional masked autoencoder and global response normalization.
\[
\textbf{Reference:}\quad
\emph{VGG}\rightarrow
\emph{ResNet}\rightarrow
\emph{ResNeXt}\rightarrow
\emph{ConvNeXt}\rightarrow
\emph{ConvNeXt V2}.
\]

\begin{table}[h]
\centering
\small
\caption{Qualitative comparison of lineage-search results for a ConvNeXt-V2-centered ConvNet lineage.}
\label{tab:case-study-resnet-lineage}
\begin{tabularx}{\linewidth}{l X}
\toprule
\textbf{Method} & \textbf{Retrieved lineage} \\
\midrule
Reference &
\emph{VGG} $\rightarrow$ \emph{ResNet} $\rightarrow$ \emph{ResNeXt}
$\rightarrow$ \emph{ConvNeXt} $\rightarrow$ \emph{ConvNeXt V2} \\
SGT-MCTS &
\emph{Caffe} $\rightarrow$ \emph{VGG} $\rightarrow$ \emph{ResNet}
$\rightarrow$ \emph{ResNeXt} $\rightarrow$ \emph{ConvNeXt}
$\rightarrow$ \emph{ConvNeXt V2} $\rightarrow$
\emph{Robust Context-Aware Object Recognition} \\
Beam@10 &
\emph{Speech Recognition with Deep RNNs} $\rightarrow$ \emph{LSTM ASR}
$\rightarrow$ \emph{Highway Networks} $\rightarrow$ \emph{ResNet}
$\rightarrow$ \emph{ResNeXt} $\rightarrow$ \emph{ConvNeXt}
$\rightarrow$ \emph{ConvNeXt V2} \\
RW@100 &
\emph{Learning to Forget} $\rightarrow$ \emph{Highway Networks}
$\rightarrow$ \emph{ResNet} $\rightarrow$ \emph{Shake-Shake}
$\rightarrow$ \emph{Decoupled Weight Decay} $\rightarrow$
\emph{ConvNeXt} $\rightarrow$ \emph{ConvNeXt V2} \\
\bottomrule
\end{tabularx}
\end{table}

SGT-MCTS recovers the complete reference subsequence in the correct order. The extra \emph{Caffe} node provides implementation context before \emph{VGG},
and the final node is a downstream user of \emph{ConvNeXt V2}; neither disrupts the main residual ConvNet backbone. Beam search recovers the later segment
from \emph{ResNet} to \emph{ConvNeXt V2}, but misses the \emph{VGG} depth-scaling step and enters the chain through an RNN/Highway-Network branch. Random
walk recovers only a shorter local subsequence around \emph{ResNet}, \emph{ConvNeXt}, and \emph{ConvNeXt V2}, while skipping both \emph{VGG} and
\emph{ResNeXt}. This case shows the value of MCTS-style exploration in a high-branching architecture graph: by keeping multiple plausible ancestors alive,
SGT-MCTS preserves the ordered VGG--ResNet--ResNeXt--ConvNeXt backbone more faithfully than the greedy and random baselines.

\section{Limitations and Broader Impact}
\label{app:limitations}

\paragraph{Extraction.}
Phase-1 edge-type classification accuracy ranges from $70.4\%$ (production model) to $93.0\%$ (audit model), reflecting the inherent difficulty of distinguishing fine-grained causal relations such as \emph{extends} versus \emph{improves}. Downstream operators treat edge types as routing rather than ground truth, so localized misclassifications shift ranking quality without breaking the operator interface. The 14-axis bottleneck taxonomy $\mathcal{D}$ is fixed at release, and emerging dimensions are mapped to the closest existing axis until a future taxonomy revision. Substring-based alias resolution favors precision over coverage and uses a manually maintained negative list to handle truly ambiguous names.

\paragraph{Algorithmic.}
The temporal coherence function $\mathrm{TC}$ is calibrated on post-2015 AI literature, so fields with markedly different research cadences would require recalibration. The evaluator's zero-trainable-parameter design is a deliberate commitment to deterministic reproducibility, trading potential accuracy gains from a learned scorer for full auditability. The four generation strategies cover the most frequent topological moves we observed; less common modes such as theoretical unification of two previously distinct frameworks are natural extensions.

\paragraph{Broader impact.}
Intern-Atlas is a research-infrastructure artifact, and its risks are largely indirect. A downstream agent consuming the graph could amplify existing citation biases when allocating credit across methodological traditions. We mitigate this through (i) verbatim-grounded evidence on every non-background edge so that consumers can audit provenance, (ii) public release of the reliability audit alongside the graph, and (iii) a stated commitment that the artifact is not to be used for author- or institution-level ranking. We see no plausible pathway to direct harm from the artifact itself.

\end{document}